\documentclass[11pt,letterpaper]{article}
\usepackage{acl2015}
\usepackage{times}
\usepackage{url}
\usepackage{latexsym}
\usepackage{graphicx}
\usepackage{amsmath}
\usepackage{amsthm}
\usepackage{multirow}
\usepackage{mathtools}

\title{Bi-directional LSTM Recurrent Neural Network for Chinese Word Segmentation}
\author{Yushi Yao\\
	    Shanghai Jiaotong University\\
	    800 Dongchuan Road\\
	    Minhang District, Shanghai, China\\
	    {\tt yys12345@sjtu.edu.cn}
	  \And
	Zheng Huang\\
  	Shanghai Jiaotong University\\
  	800 Dongchuan Road\\
  	Minhang District,Shanghai , China\\
  {\tt huangzheng@sjtu.edu.cn}}

\date{}

\begin{document}
\maketitle
\begin{abstract}
	Recurrent neural network(RNN) has been broadly applied to natural language processing(NLP) problems. This kind of neural network is designed for modeling sequential data and has been testified to be quite efficient in sequential tagging tasks. In this paper, we propose to use bi-directional RNN with long short-term memory(LSTM) units for Chinese word segmentation, which is a crucial preprocess task for modeling Chinese sentences and articles. Classical methods focus on designing and combining hand-craft features from context, whereas bi-directional LSTM network(BLSTM) does not need any prior knowledge or pre-designing, and it is expert in keeping the contextual information in both directions. Experiment result shows that our approach gets state-of-the-art performance in word segmentation on both traditional Chinese datasets and simplified Chinese datasets.
\end{abstract}

\section{Introduction} 
	With the rapid development of deep learning, neural networks start to show its great capability in NLP tasks[Auli {\em et al.}, 2013] and recent research revealed that recurrent neural networks(RNN) significantly outperforms popular statistical algorithms like Hidden Markov Model(HMM)[Zhang {\em el al.}, 2003],  CRF(conditional random field)[Peng {\em et al.}, 2004] and neural probabilistic models[Bengio {\em et al.}, 2003]. 
	
	As a special kind of RNN,  LSTM neural networks[Hochreiter {\em et al.}, 1997] is verified to be efficient in modeling sequential data like speech and text [Sundermeyer {\em et al.}, 2015]. More over, BLSTM neural network[Schuster {\em et al.}, 1997], which is derived from LSTM network, has advantages in memorizing information for long periods in both directions, making great improvement in linguistic computation. Sundermeyer et al. analyzed LSTM neural network by modeling English and French[Sundermeyer {\em et al.}, 2012]. Wang et al. used bi-directional LSTM into POS tagging, chunking and NER tasks and internal representations are learnt from unlabeled text for all tasks[Wang {\em et al.}, 2015]. Huang et al. combined LSTM with CRF and verified the efficiency and robustness of their model in sequential tagging[Huang {\em et al.}, 2015]. Ling et al focus on constructing compact vector representations of words with bi-directional LSTM, which yield state-of-the-art performance in contrast to other word-to-vector algorithms like CBOW and skip-n-gram[Ling {\em et al.}, 2015]. The study that is close to ours is Chen et al, which introduced LSTM neural network into Chinese word segmentation[Chen {\em et al.}, 2015], while LSTM can just memorize the past contextual information from the context. Due to the complicated and changeable structure of Chinese sentence, it's intuitive that both future and past information need to be considered when training segmentation model.
		
	In this study, in order to improve the performance of the Chinese word segmentation, we applied bi-directional LSTM networks to word segmentation task, we also constructed higher-level features of characters with BLSTM network and our contributions can be listed as follows:\\
	\indent1) Our work is the first to apply and improve bi-directional LSTM network to Chinese word segmentation benchmark data sets\\
	\indent2) The training framework can be regarded as an integration of generating embeddings and tagging characters and it does not need any external datasets
	 
	The paper is organized as follows. Section 2 describes basic idea and architecture of BLSTM network. Next, we introduce our training framework based on BLSTM network in Section 3. Section 4 details our experiments on Chinese dataset and summarizes our experimental results with previous research. Finally, in Section 5 we summarize key conclusions.
	
\section{BLSTM network architecture}
	BLSTM neural network is similar to LSTM network in structure because both of them are constructed with LSTM units[Schuster {\em et al.}, 1997]. The special unit of this network is capable of learning long-term dependencies without keeping redundant context information. They work tremendously well on sequential modeling problems, and are now widely used in NLP tasks.

\subsection{LSTM unit}
	The basic structure of LSTM memory unit is composed of three essential gates and a cell state. 
	\begin{figure}[htb]
  	\centering
  	\includegraphics[width=2.7in]{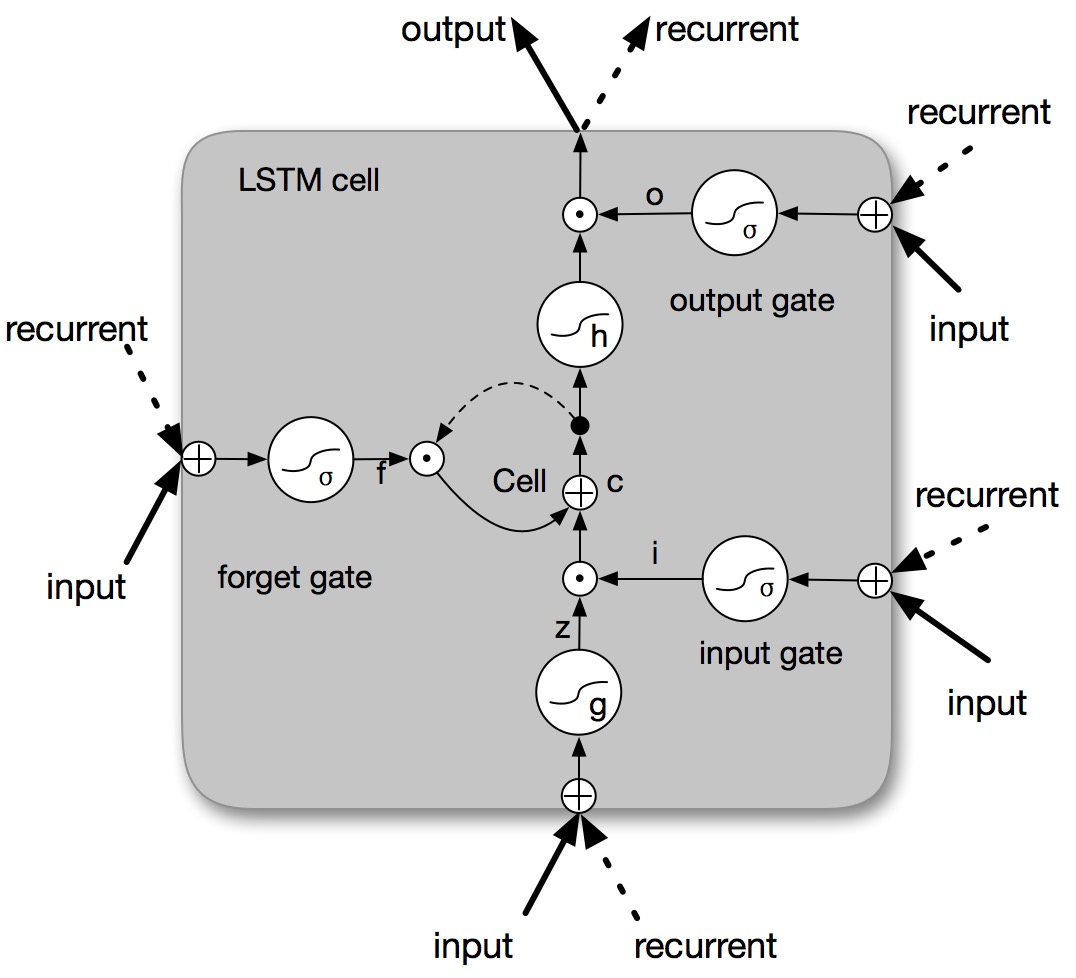}
 	 \caption{Schematic of LSTM unit}
  	\end{figure}
	
	As shown in Figure 1, the memory cell contains the information it memorized at time $t$, the state of the memory cell is bound up together with three gates, the input vector of each gate is composed of input part and recurrent part. Forget gate controls what to abandon from the last moment, input gate decides what new information will be stored in the cell state, the output gate decides which part of the cell state will be output and the recurrent part will be updated by current cell state and fed into next iteration[Hochreiter {\em et al.}, 1997].
	
	The formal formulas for updating each gate and cell state are defined as follows:
	\begin{align}
	&z^t = g(W_z x^t + R_z y^{t-1} + b_z)\\
	&i^t = \sigma(W_ix^t + R_iy^{t-1} + p_i \odot c^{t-1} + b_t)\\
	&f^t = \sigma(W_fx^t + R_fy^{t-1} + p_f \odot c^{t-1} + b_f)\\
	&c^t = i^t\odot z^t + f^t \odot c^{t-1}\\
	&o^t = \sigma(W_ox^t + R_oy^{t-1} + p_o \odot c^{t} + b_o)\\
	&y^t = o^t \odot h(c^t)
	\end{align}
	
	Here $x^t \in R^d$ and $y^t \in R^d$ are input and output vector of the unit at time$t$, $W_k(k = z,i,f,o)$ and $R_k(k = z,i,f,o)$ are weight matrices for input part and recurrent part of different gates, $b_k (k = z,i,f,o)$ denotes bias vector and the functions $\sigma$, $g$ and $h$ are non-linear functions such as sigmoid or tanh, $\odot$ means point-wise calculation of two vectors. For completeness, we add $p_k (k= i, o, f)$ to the formulas, which denote peephole connection and is mostly used in LSTM variants. 

\subsection{BLSTM Network}
	BLSTM network is designed to capture information of sequential dataset and maintain contextual features from past and future. Different from LSTM network, BLSTM network has two parallel layers propagating in two directions, the forward and backward pass of each layer are carried out in similar way of regular neural networks, these two layers memorize the information of sentences from both directions.[Schuster {\em et al.}, 1997]	
	
	Since there are two LSTM layers in our network, the vector formula should be also adjusted. 
	\begin{align}
	&{h_f}_t = H(W_{x h_f}x_t + W_{h_f h_f}{h_f}_{t-1} + b_{h_f}) \\
	&{h_b}_t = H(W_{x h_b}x_t + W_{h_b h_b}{h_b}_{t-1} + b_{h_b})
	\end{align}
	$h_f \in R^d$ and $h_b \in R^d$ denotes the output vector of forward layer and backward layer respectively, different from former research, the final output in our work $y_t = [h_{f_t}, h_{b_t}]$ is the concatenation of these two parts, which means $y_t \in R^{2d}$.  We define the combination of forward and backward layers as a single BLSTM layer.  

\section{Training Method}
	In order to convert the segmentation problem into a tagging problem, we assign a label for each character to indicate the segmentation. There are four kinds of labels: \textit{B, M, E, S}, corresponding to the beginning, middle, end of a word, and a single-character word, respectively.
%
\subsection{Training framework}
	The basic procedure of language modeling in our study is shown in Figure 2.
	
	Each character has an id which is defined in a lookup dictionary, the dictionary is constructed by collecting unique characters in the training set. Instead of one-hot representation, the characters are projected into a d-dimension space and initialized as dense vectors $v \in R^d$, we regard this initialization step as constructing embeddings for characters. Every embedding is stored in a matrix $M \in R^{d \times |C|}$ and can be retrieved by its character id. As embeddings are efficient in describing word-level features[Miklov {\em et al.}, 2013], we hope character-level embeddings can also achieve good performance in CWS. 
	\begin{figure}[htb]
  	\centering
  	\includegraphics[width=2.8in]{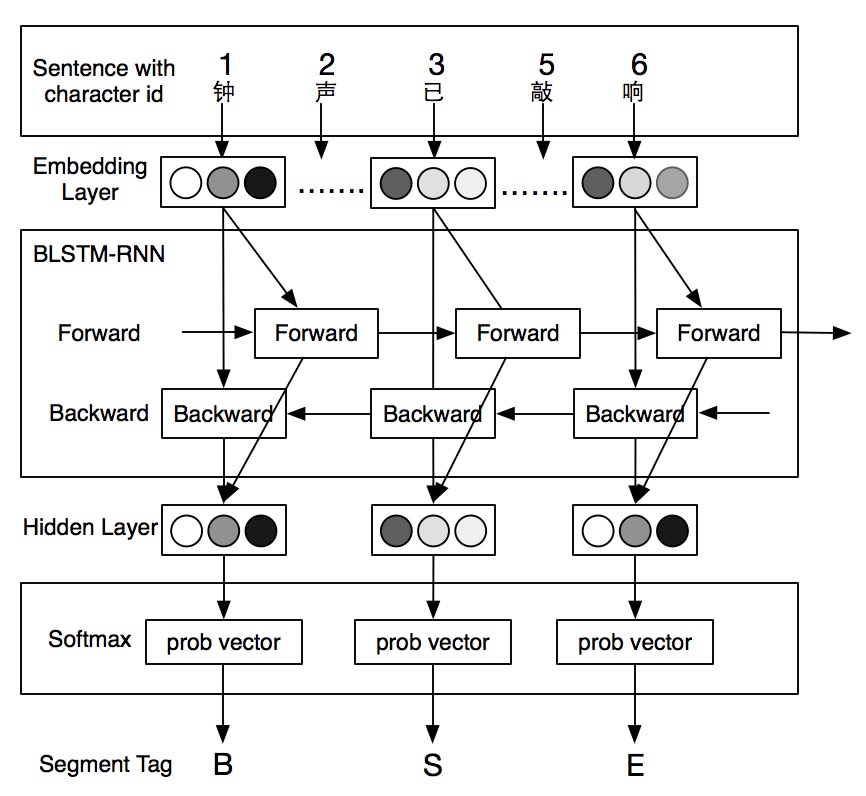}
 	\caption{Illustration of BLSTM Language Modeling for CWS}
  	\end{figure}
	Then he embeddings are fed into BLSTM network and the final output of BLSTM network is finally passed to a hidden layer and the softmax layer determines the tag with maximum probability of the character.
	
\subsection{Model variant}
	
	In order to further improve the structure of BLSTM network, we stack BLSTM layers based on the method of constructing RNN[Pascanu {\em et al.}, 2013]. expecting to extract contextual features in higher level.
	
	However, the output of a BLSTM layer is in double size of the input vector since it is composed of two LSTM layers, and its dimension will expand dramatically when the network goes deeper, here we use a transformation matrix to compress the dimension of output vectors, and keep it the same size with input vectors. 
	$$ v_{tran} = W_{tran} \times v_{o} \eqno{(9)}$$
	Assume that the output vector of BLSTM layer $v_o \in R^{2d}$, the transformation matrix $W_{tran} \in R^{d \times 2d}$ convert the vectors into lower dimension and thus the output of each BLSTM layer is kept the same dimension. As our BLSTM network gets more and more complicated, the number of parameters grows rapidly and we used dropout during training in order to avoid overfitting[Srivastava {\em et al.}, 2014]. 

\section{Experiments}
	\textbf{Setup} The dataset we used for evaluating our model on word segmentation is from Backoff 2005, which contains benchmark datasets for both simplified Chinese(PKU and MSRA) and traditional Chinese(AS and HK).
	All the models are trained on NVIDIA GTX Geforce 970, it took about 16 to 17 hours to train a model on GPU while more than 4 days to train on CPU, in contrast. We also changed batch size during our training process because of the limit of GPU memory. 
	
	
	\noindent \textbf{Results} In this section, we will state the procedure of our experiments and how we get the model with the best performance, and we will also compare the performance of our network with state-of-the-art approaches. 

	\begin{table*}[tb]
	\begin{center}
	\begin{tabular}{*{13}{c}}
	\hline
	\multirow{2}*{Models}
	 & \multicolumn{3}{c}{PKU} & \multicolumn{3}{c}{MSRA} 
	 & \multicolumn{3}{c}{AS} & \multicolumn{3}{c}{HKCityU}\\\cline{2-13}
	 & P & R & F & P & R & F & P & R & F & P & R & F\\\hline
	 Bi-RNN & 94.2 & 92.5 & 93.3 & 95.7 & 94.8 & 95.2 & 96.1 & 96.4 & 96.2 & 96.8 & 95.9 & 96.3\\
	 LSTM* & 95.3 & 94.6 & 94.9 & 96.1 & 95.3 & 95.7 & 94.2 & 93.2 & 93.7 & 97.2 & 96.6 & 96.9\\
  	 BLSTM* & 96.5 & 95.3 & 95.9 & 96.6 & 97.1 & 96.9 & 97.3 & 96.9 & 97.1 & 97.4 & 97.2 & 97.3\\
	 BLSTM2* & 96.6 & 95.9 & 96.2 & 97.3 & 97.1 & 97.2 & 97.9 & 97.5 & 97.7 & 97.5 & \bf97.4 & 97.4\\
	 BLSTM3* & \bf96.8 & \bf96.3 & \bf96.5 & \bf97.4 & \bf97.3 & \bf97.3 & \bf98.0& \bf97.6 & \bf97.8 & \bf97.7 & 97.3 & \bf97.5\\\hline
	\end{tabular}
	\end{center}
	\caption{\label{font-table} Performance of our models on four test sets, add simple RNN and LSTM network for comparison. BLSTM2 means a bi-directional LSTM network with two layer, and BLSTM3 means a bi-directional LSTM network with three BLSTM layers,  * denotes that the model is trained with dropout}
	\end{table*}
	Since the dimension of embedding vector will directly influence the number of parameters and model complexity, we conducted related experiments on HKCityU dataset, and try to get a suitable size of embedding first. 
	
	
	Table 2 illustrate the performance of our model on HKCityU dataset with embedding vectors of different dimensions. When the dimension grows higher, the error rate becomes bigger and unstable. We can conclude that the model gets the best performance when the embedding dimension is 200, which also indicates that an efficient representation of Chinese words should not be too long.  
	
	\begin{table}[!h]
  	\begin{center}
 	 \begin{tabular}{cccc}
  	 \hline Embedding Size & P & R & F\\ \hline
  	 100 & 92.2 & 91.6 & 91.9\\
	 128 & 92.6 & 92.1 & 92.3 \\
 	 200 & \bf 94.1 & \bf 93.5 & \bf 93.8 \\
  	 300 & 90.6 & 90.2 & 90.4\\
  	\hline
  	\end{tabular}
  	\end{center}
  	\caption{\label{font-table} performance of BLSTM networks with different embedding dimensions}
  	\end{table}
	
	Secondly, we tested the performance of our model with different number of BLSTM layers. 
	
	Table 1 shows that as we stack more BLSTM layers, the performance gets slight improvement, while adding layers becomes not so effective when the number of BLSTM layers exceeds three, which also takes quite long time to train. The result shows that LSTM units become less effective in higher level layers so we believe that there is no need to build very deep network for extracting contextual information.
	 
	 \begin{table}[!h]
  	 \begin{center}
 	 \begin{tabular}{cccc}
  	 \hline Models & PKU & MSRA & CityU\\ \hline
  	 (Zhao et al., 2006) & - & - & \bf97.7 \\
	 (Sun and Xu, 2011) & 95.1 & 97.2 & -\\
 	 (Zhang et al., 2013) &  96.1 & 97.4 & -\\
  	 (Chen et al., 2015) & \bf96.5 & 97.4 & - \\
  	 Ours & \bf96.5 & \bf97.6 & 97.5\\
  	\hline
  	\end{tabular}
  	\end{center}
  	\caption{\label{font-table} Comparison of our model with previous research}
  	\end{table}

	Table 3 lists the performances of our model as well as previous research. (Zhao et al., 2006) is a CRF model with rich feature template, (Sun and Xu, 2011) improved supervised word segmentation by exploiting features of unlabeled data and the system of (Zhang {\em et al.}, 2013) applied semi-supervised approach to extract representations of label distributions from unlabeled and labeled datasets[Zhang {\em et al.}, 2013]. Nevertheless, all the models or systems above focus on feature engineering, while our approach do not depend on any predesigned features thanks to the strong ability of BLSTM network in automatic feature learning.
	
	Our model achieved competitive performance compared to the work of Chen's, which also used character embeddings but applied LSTM network to CWS, and the results suggest that BLSTM may get better performance than LSTM on segmentation, and indicates that both past and future information should be taken into account for segmentation task.
	
\section{Conclusions}
	In this paper, we propose to use bi-direction LSTM neural network to train the model for Chinese word segmentation, BLSTM network is quite efficient for sequential tagging task. The model learn to extract discriminative character-level features automatically and it do not require any hand-craft features for segmentation or prior knowledge. Experiments conducted on SIGHAN Backoff 2005 datasets show that our model has good performance and generalization on both simplified Chinese and traditional Chinese. Our results suggest that deep neural networks work well on segmentation tasks and BLSTM networks with word embedding is an effective tagging solution and worth further exploration.


\end{document}